\documentclass[sigconf]{acmart}

\usepackage{multirow} 
\usepackage{graphicx}
\usepackage{ragged2e}
\usepackage{array}
\usepackage{longtable}
\usepackage{rotating}
\usepackage[justification=centering]{caption}

\AtBeginDocument{%
  }

\setcopyright{acmcopyright}
\copyrightyear{2023}
\acmYear{2023}
\acmDOI{XXXXXXX.XXXXXXX}

\acmConference[Conference acronym 'XX]{Make sure to enter the correct
  conference title from your rights confirmation emai}{June 03--05,
  2018}{Woodstock, NY}
\acmPrice{15.00}
\acmISBN{978-1-4503-XXXX-X/18/06}

\begin{document}

\title{DSformer: A Double Sampling Transformer for Multivariate Time Series Long-term Prediction}



\author{Chengqing Yu}
\orcid{1234-5678-9012}
\affiliation{%
  \institution{Institute of Computing Technology,\\Chinese Academy of Sciences}
  \country{University of Chinese Academy of Sciences}
  \postcode{100190}
}
\email{yuchengqing22b@ict.ac.cn}

\author{Fei Wang}
\authornote{Corresponding author.}
\affiliation{%
  \institution{Institute of Computing Technology,}
  \country{Chinese Academy of Sciences}
  \postcode{100190}
}
\email{wangfei@ict.ac.cn}

\author{Zezhi Shao}
\affiliation{%
  \institution{Institute of Computing Technology,\\Chinese Academy of Sciences}
  \country{University of Chinese Academy of Sciences}
  \postcode{100190}
}
\email{Shaozezhi19b@ict.ac.cn}

\author{Tao Sun}
\affiliation{%
  \institution{Institute of Computing Technology,}
  \country{Chinese Academy of Sciences}
  \postcode{100190}
}
\email{suntao@ict.ac.cn}

\author{Lin Wu}
\affiliation{%
  \institution{Institute of Computing Technology,}
  \country{Chinese Academy of Sciences}
  \postcode{100190}
}
\email{wulin@ict.ac.cn}

\author{Yongjun Xu}
\affiliation{%
  \institution{Institute of Computing Technology,}
  \country{Chinese Academy of Sciences}
  \postcode{100190}
}
\email{xyj@ict.ac.cn}

\renewcommand{\shortauthors}{Yu et al.}

\begin{abstract}
  Multivariate time series long-term prediction, which aims to predict the change of data in a long time, can provide references for decision-making. Although transformer-based models have made progress in this field, they usually do not make full use of three features of multivariate time series: global information, local information, and variables correlation. To effectively mine the above three features and establish a high-precision prediction model, we propose a double sampling transformer (DSformer), which consists of the double sampling (DS) block and the temporal variable attention (TVA) block. Firstly, the DS block employs down sampling and piecewise sampling to transform the original series into feature vectors that focus on global information and local information respectively. Then, TVA block uses temporal attention and variable attention to mine these feature vectors from different dimensions and extract key information. Finally, based on a parallel structure, DSformer uses multiple TVA blocks to mine and integrate different features obtained from DS blocks respectively. The integrated feature information is passed to the generative decoder based on a multi-layer perceptron to realize multivariate time series long-term prediction. Experimental results on nine real-world datasets show that DSformer can outperform eight existing baselines. 
\end{abstract}



\received{20 February 2007}
\received[revised]{12 March 2009}
\received[accepted]{5 June 2009}

\maketitle

\section{Introduction}

Multivariate time series prediction is widely used in our life, such as weather \cite{RN917}, energy \cite{RN908}, economics \cite{RN919}, environment \cite{RN918}, traffic \cite{RN25} and other fields \cite{mengchang1} \cite{liang2023message}  \cite{RN26}. Specially, multivariate time series long-term prediction can help people understand the changing trend of data for a long time in the future, which provides important references for decision-making \cite{RN859} \cite{RN51}. Therefore, multivariate time series long-term prediction has always been a hot topic in academia \cite{mengchang2} and industry \cite{RN921}.

Multivariate time series is composed of multiple time series with correlations \cite{RN920}. And these correlated time series usually fluctuate and change over time \cite{RN907}. As a special sequence form different from natural language, researchers usually need to analyze the time series context relation \cite{RN916}  and variable correlation \cite{RN913} of multivariate time series data to achieve long-term prediction. At present, transformer-based models are widely studied in this field because of their powerful sequential modeling and context relation analysis capabilities \cite{RN887}. However, these models do not make full use of three features of multivariate long sequence time series. And based on Figure 1, we introduce these features next:

\begin{figure}
  \centering
  \includegraphics[width=\linewidth]{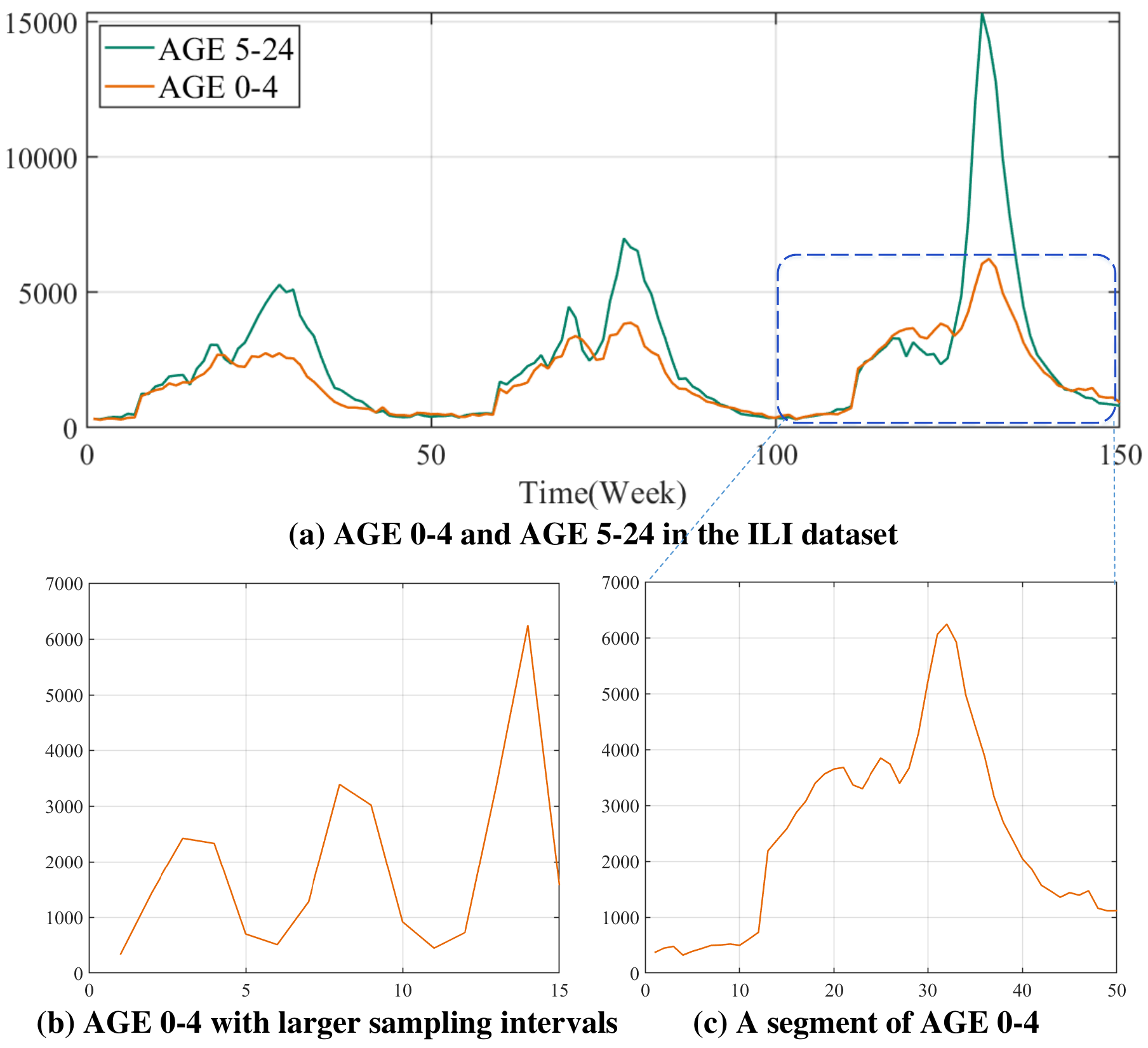}
  \caption{Examples of the multivariate time series in ILI dataset. (a) Time series of variable AGE 5-24 and variable AGE 0-4 in the ILI dataset. (b) AGE 0-4 time series with larger sampling intervals. (c) A segment of AGE 0-4 time series.}
  \Description{Examples of multivariate time series data.}
\end{figure}

\begin{itemize}
\item \textbf{Variable correlation:} As shown in Figure 1 (a), two correlated time series show similar change patterns over time. If the model can find the relationship between these two time series, that is, variable correlation, it can mine more information and improve the modeling effect.
\item \textbf{Global information:} When the sampling frequency of the AGE 0-4 data in Figure 1 (a) is increased, the raw data can be transformed into the time series shown in Figure 1 (b). By observing Figure 1 (b), we find that the processed data shows seasonality on global. In other words, the proposed data is composed of multiple similar segments. If the model finds this global information, it can predict the overall future changes of the data.
\item \textbf{Local information:} As shown in Figure 1 (c), when we focus on observing one part of three similar segments of AGE 0-4 data in Figure 1 (a), we can capture more detailed local information than Figure 1 (b). Therefore, if the model can combine this information with the above global information, it won't lose local details in the process of modeling. 
\end{itemize}

Based on the above analysis, if we can effectively use these three features (global information, local information and variables correlation) of multivariate long sequence time series, the model can be more suitable for long-term prediction. However, we need to address the following technical challenges: (1) How do we make our model observe these three features of the original data? (2) How to effectively integrate these features to achieve multivariate time series long-term prediction? 

To mine the above three features of the multivariate long sequence time series, we propose a double sampling (DS) block and a temporal variable attention (TVA) block, which can mine these features from the following aspects: (1) The DS block uses two components (down-sampling method and piecewise sampling method) to process the raw data. The down-sampling method obtains the feature vector by extracting the original data with a larger sampling interval, as shown in Figure 1 (b). Observing the data with larger sampling intervals can reduce the influence of local noise and obtain more global information. And the piecewise sampling method obtains the local time series by splitting the original data proportionally, as shown in Figure 1 (c). Observing a continuous segment of a long sequence can enhance the utilization of local information. After processing by the DS block, we can obtain two feature vectors containing global information or local information respectively. (2) The TVA block uses a parallel modeling structure to combine temporal attention and variable attention, and mine above feature vectors. Specifically, temporal attention analyzes context relation and captures the information from temporal dimension (global information or local information). And variable attention focuses on analyzing the variable correlation. Besides, different from the traditional idea of stacking multiple layers, we use temporal attention and variable attention to mine feature vectors respectively, and then integrate the extracted information. Based on the above ideas, the TVA block can mine and integrate temporal information (global information or local information) and variable correlation. Then, we need to further integrate above three key features. 

To further mine and integrate above three key features (local information, global information and variable correlation), we still use the idea of parallel modeling to mine and integrate the two feature vectors obtained by DS block. Specifically, multiple TVA blocks are used to model feature vectors obtained by DS block separately and integrate the processed features. Firstly, two TVA blocks are used to separately mine two different feature vectors obtained by DS block. And the TVA block introduce the variable correlation while mining the global information or local information owned by above two feature vectors respectively. Then, we use a TVA block to combine the above feature vectors and obtain the feature vector that integrates these key information. Finally, the integrated feature vector is transmitted to the generative decoder for prediction modeling. Based on the above blocks and modeling steps, we finally proposed the double sampling transformer (DSformer). \textbf{In general, the main contributions of this paper are shown as follows}:

\begin{itemize}
\item We propose a novel model for multivariate time series long-term prediction, which is called DSformer. It learns and integrates global information, local information and variables correlation of multivariate time series. 
\item We design a double sampling block to preprocess the original data and help the model mine the global information and local information. Besides, we propose a temporal variable attention block to mine the data from the temporal dimension and variable dimension. These two blocks are combined by a parallel structure for information integration. 
\item We conduct comparative experiments on nine real world data sets. The results demonstrate that DSformer can outperform eight existing SOTA models.  
\end{itemize}

\section{Related Work}
\subsection{Deep learning based methods}
At present, deep learning has been widely studied in the field of multivariate time series long-term prediction \cite{RN841}. As one of the most classical deep learning algorithms in time series prediction, recurrent Neural Network (RNN) \cite{RN735} has been widely studied. As the most classical variant of RNN, the long short-term memory network (LSTM) \cite{RN275} and the gated recurrent unit (GRU) \cite{RN761} have made progress in the field of time series prediction. Compared with RNN, LSTM and GRU effectively solve the gradient problem and improve their prediction accuracy \cite{liang2023knowledge}. In addition to RNN-based models, the convolutional neural network (CNN) \cite{RN839} based models have also been proven to have effects in the field of multivariate time series long-term prediction. For example, Temporal Convolutional Network (TCN) \cite{RN845} improves the ability of sequence modeling by introducing Dilated Causal Convolutions and residual connections. Besides, with the improvement of computer performance, the idea of fusing different network structures is constantly proposed \cite{RN774}. LSTMa \cite{RN910} improved the ability of the model to mine temporal information by combining LSTM and attention mechanism. Besides, by effectively combining LSTM, CNN and attention mechanism, LSTNet \cite{RN857} achieved better results than traditional methods in multivariate time series long-term prediction. However, the above models have limitations in mining the key context information of long sequence and the correlation of different variables, which limits their performance.

\subsection{Transformer based methods}
Due to its excellent series modeling capabilities, Transformer variants have seen rapid growth in multivariate time series long-term prediction \cite{zerveas2021transformer}. Li et al. \cite{RN906} used the convolutional self-attention mechanism to improve the sequence modeling ability of the traditional transformer and proposed the LogSparse transformer (LogTrans). Kitaev et al. \cite{RN911} combined Locality sensitive hashing attention with reversible residual layers to improve the ability to analyze long-term dependencies and proposed Reformer. Zhou et al. \cite{RN742} proposed Informer by introducing a ProbSparse self-attention mechanism and the generative decoder. Liu et al. \cite{RN909} proposed Pyraformer, which introduces the pyramidal attention module and multi-resolution modeling approach. The above models focus on optimizing the ability of attention to analyze the long-term dependence, but they do not fully analyze the characteristics of time series. Different from the above methods, Autoformer \cite{RN886} introduces autoregressive attention and deep decomposition structure to realize long-term prediction of time series. The deep learning decomposition structure improves the ability of the model to analyze trends and seasons. On this basis, FEDformer \cite{RN912} and TDformer \cite{RN905} introduce the deep decomposition framework and Fourier Attention to realize the long-term prediction of time series. These methods improve the ability to mine time series context relation by introducing trend and seasonal modeling. However, the decomposition method transform raw data into fixed forms based on expert experience, which limits the ability of the model to mine the data itself. At present, to strengthen the model's ability to mine global information from raw data, Patch TST \cite{RN20} and Crossformer \cite{RN19} adopt the idea of patch modeling. In addition, Patch TST and Crossformer respectively adopt channel independence and two-stage attention to realize multivariable modeling. Due to the mining of more data features, Patch TST and Crossformer can achieve better capabilities than the transformer variants mentioned above. However, Patch TST ignores the correlation between different variables, and Crossformer ignores the role of local information. In general, the existing models do not make full use of the key features of multivariate long sequence time series, which limits their performance.

\section{Methodology}
\subsection{Preliminaries}
In this section, we introduce the basic definition of multivariate time series and multivariate time series long-term prediction. 

\noindent\textbf{Multivariate time series.} The multivariate time series is a data form composed of multiple sequences that change over time \cite{RN62}. The representation of multivariate time series can usually be defined as a tensor $X\in R^{N*T}$ \cite{choi2022graph}. $N$ is the number of variables. $T$ is the length of the time step.

\noindent\textbf{Multivariate time series long-term prediction.} Given a historical sequence $X\in R^{N*H}$ from $H$ time steps in history, the model can predict the value $Y\in R^{N*L}$ of the nearest $L$ time steps in the future \cite{shao2023hutformer}. The main purpose of multivariate time series long-term prediction is to establish the mapping relationship between input $X\in R^{N*T}$ and label $Y\in R^{N*L}$ \cite{RN915}.

\subsection{Overall framework of the proposed model}
The overall framework of the DSformer is given in  Figure 2. And it can be found that DSformer contains two important component: the double sampling block and the temporal variable attention block. And DSformer combines a double sampling block and three temporal variable attention blocks to mine three features and fully perform information integration. In this section, we intuitively discuss each block of DSformer and its parallel structure.

 \begin{figure}[h]
  \centering
  \includegraphics[width=\linewidth]{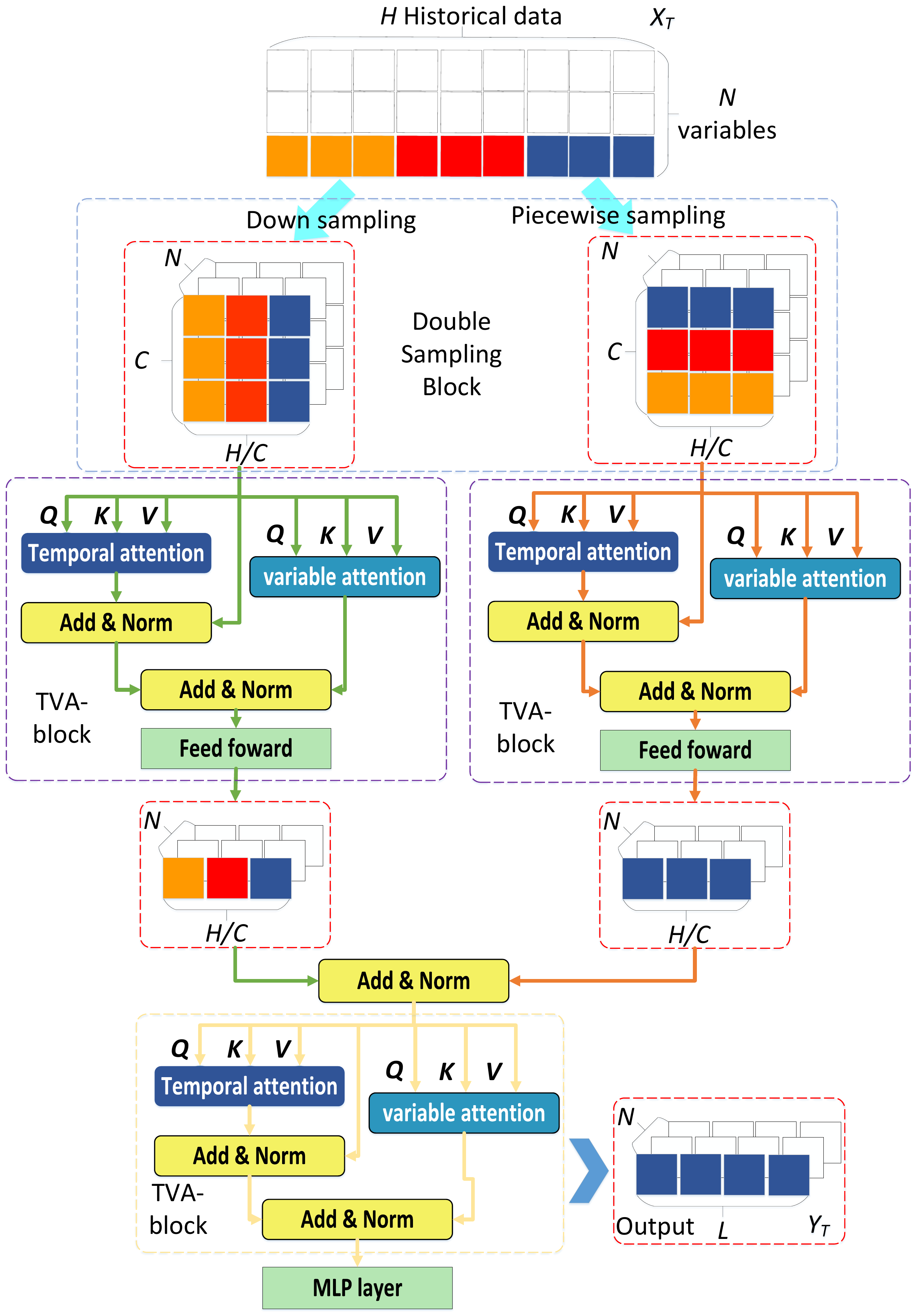}
  \caption{Overall framework of the proposed DSformer, the DS block and the TVA block.}
  \Description{Overall framework of the proposed DSformer.}
\end{figure}

Firstly, we discuss the DS block, which uses the downsampling and the piecewise sampling to process the original input features respectively. The downsampling converts the original sequence into multiple subsequences with simliar length by increasing the sampling interval. The global information of subsequences with larger time intervals is more significant than that of the original sequence \cite{RN27}. The piecewise sampling can divide the long sequence into multiple contiguous fragments. Because the observation length of the time series is reduced, the model can mine the local information more intensively \cite{zhang2022less}. At the same time, to reduce the information loss caused by sampling, we connect the subsequences obtained from the sampling method and convert them into 3D tensors. 

Second, the TVA block aims to mine the 3D tensors processed by the DS block from the temproal dimension and the variable dimension. Based on the parallel structure, the TVA block enable the temporal attention and the variable attention to mine input features respectively. Different from the traditional stacked multi-layer structure, the parallel structure enables the model to mine information more centrally \cite{RN24}. Then the effective integration of temporal information and variable information is realized through addition and layer normalization.

Finally, the overall framework of DSformer also adopts parallel structure to realize feature mining and modeling. Specifically, the two different 3D tensors obtained by the DS block are mined by two TVA blocks. And then, a new TVA block is used to achieve the fusion of above two processed tensors. Therefore, DSformer can be used to mine global information, local information and variable correlation in parallel. Based on this structure, DSformer can strengthen the ability of mining features and achieving fusion.

\subsection{Double sampling block}
The double sampling block consists of two important steps: the down sampling and the piecewise sampling. Figure 3 presents a schematic of these two sampling methods. These two sampling methods transform the original 2D feature vectors $X\in R^{N*H}$ into two 3D features $X_{ds}\in R^{N*C*\frac{H}{C}}$ and $X_{ps}\in R^{N*C*\frac{H}{C}}$. The feature vector obtained by downsampling contain more global information. The feature vector obtained by piecewise sampling contains more local information. In the following, we briefly describe the proposed two sampling methods.

\textbf{Down sampling.} For a time series with length $H$, we obtain $C$ subsequences of consistent length in the same way as shown in Figure 3 (a). The subsequence obtained by the downsampling method has a larger time interval. As a special form of sequence, observing time series data with larger time intervals can obtain more intuitive global information. In addition, to avoid the information loss caused by down-sampling, we put $C$ subsequences together and obtain the feature vector $X_{ds}\in R^{N*C*\frac{H}{C}}$ for subsequent modeling.  For the $jth$ subsequence, its main constituent form is given as follows:
\begin{equation}
{X^{j}}_{ds} = [x_{j},x_{j+\frac{H}{C}},x_{j+2*\frac{H}{C}},...,x_{j+(C-1)*\frac{H}{C}}],
\end{equation}

\textbf{Piecewise sampling.}  For a time series with length $H$, we obtain $C$ subsequences of consistent length in the same way as shown in Figure 3 (b). The piecewise sampling method can transform the original time series into continuous subsequence with the same length. Each subsequence contains local information over a historical period of time. Unlike down sampling, piecewise sampling allows the model to focus more attention on local information, which usually reflects the details of local changes over a cycle. In addition, to avoid the information loss caused by piecewise sampling, we put $C$ subsequences together and obtain the feature vector $X_{ps}\in R^{N*C*\frac{H}{C}}$ for subsequent modeling. For the $jth$ subsequence, its main constituent form is given as follows:
\begin{equation}
{X^{j}}_{ps} = [x_{1+(j-1)*C},x_{2+(j-1)*C},x_{3+(j-1)*C},...,x_{j*C}],
\end{equation}

After obtaining two different feature vectors $X_{ds}\in R^{N*C*\frac{H}{C}}$ and $X_{ps}\in R^{N*C*\frac{H}{C}}$ by the DS block, we next introduce how to use TVA block to mine above feature vectors from temporal dimension and variable dimension.

\begin{figure}[h]
  \centering
  \includegraphics[width=\linewidth]{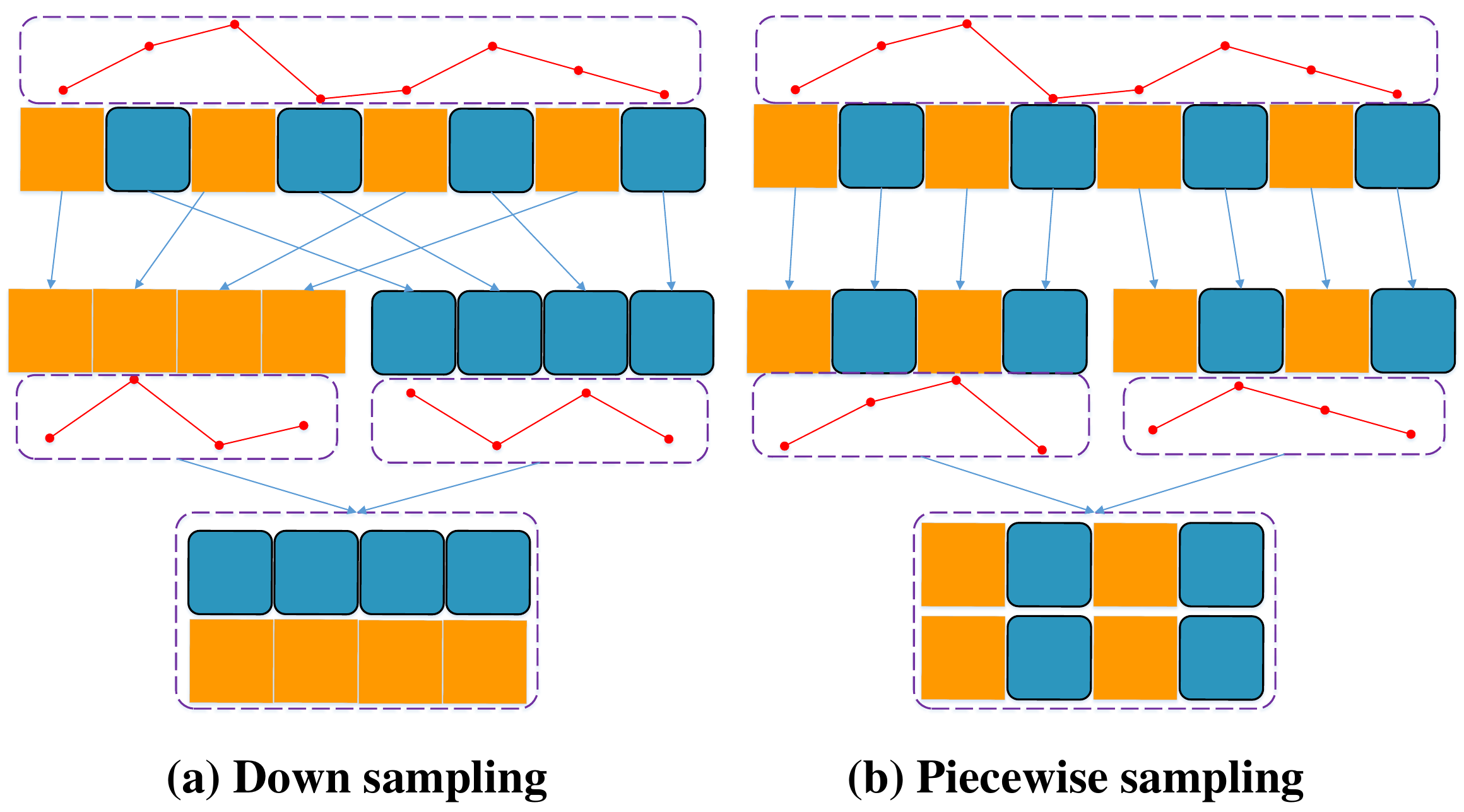}
  \caption{Schematic of the down sampling method and the piecewise sampling method.}
  \Description{Schematic of down sampling and Piecewise sampling.}
\end{figure}

\subsection{TVA block}
The proposed TVA block consists of two main components: temporal attention and variable attention. The main function of temporal attention is to mine the context information of data from the temporal dimension. The main function of variable attention is to mine the internal implicit relation between different variables. The information mined by these two components is integrated through a parallel structure. Figure 4 illustrates the detailed composition of TVA block, temporal attention and variable attention. Next, we present the modeling details of temporal attention, variable attention, and TVA blocks. 

\begin{figure}[h]
  \centering
  \includegraphics[width=\linewidth]{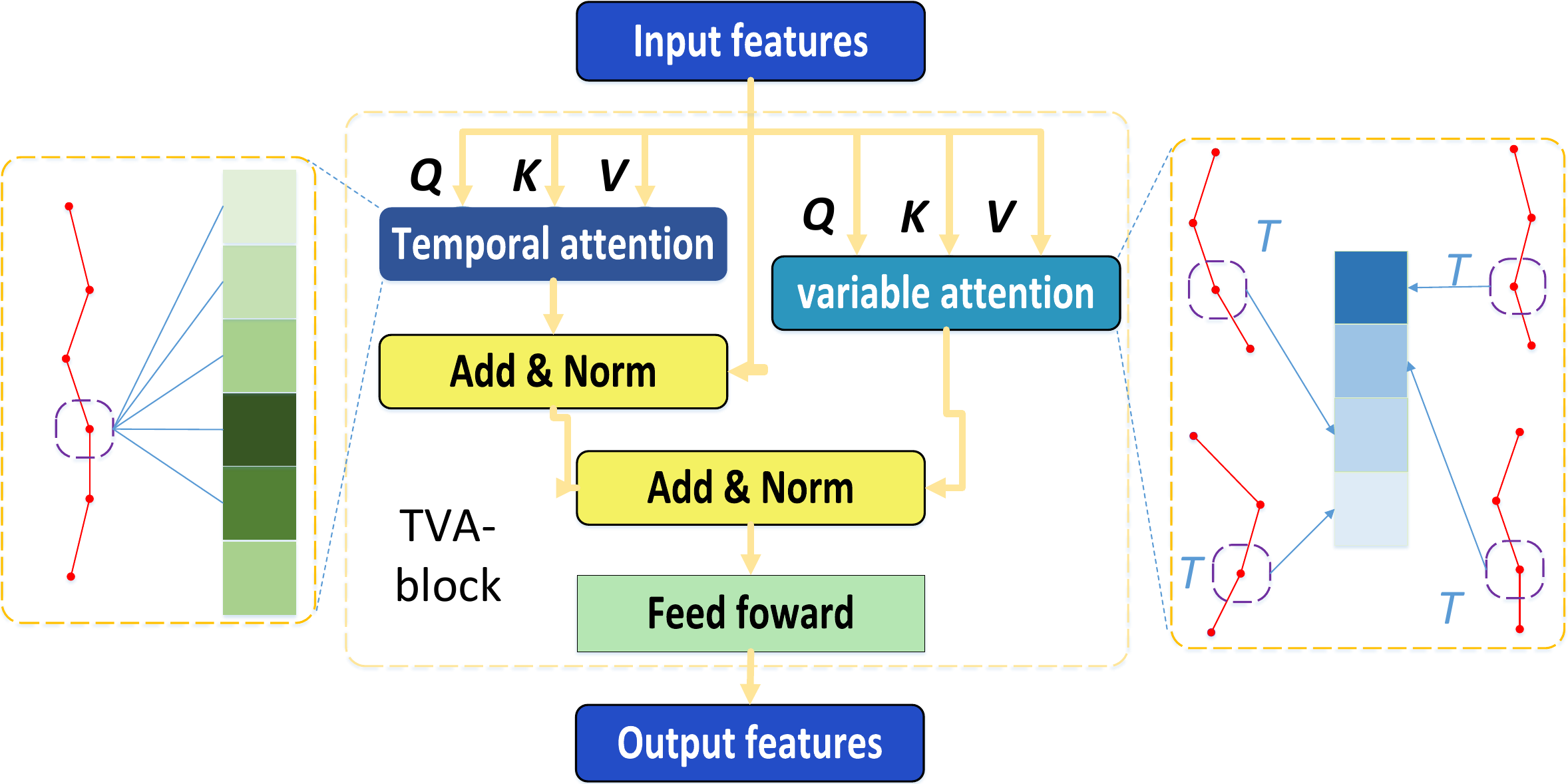}
  \caption{Schematic diagram of TVA block, temporal attention, and variable attention.}
  \Description{Schematic diagram of TVA block, temporal attention, and variable attention.}
\end{figure}

For the $X_{ds}\in R^{N*C*\frac{H}{C}}$ and $X_{ps}\in R^{N*C*\frac{H}{C}}$ obtained by the double sampling block, they are transmitted to the temporal attention and variable attention as input to the TVA block. Then, temporal attention and variable attention will process above two feature vectors in the following way:

\textbf{Temporal attention.} Temporal attention consists of three main components (multi-head attention, residual connection, and layer normalization). 

Firstly, multi-head attention is used to mine the time dimension of the input feature vector $X_{ds}\in R^{N*C*\frac{H}{C}}$ and obtain the processed feature vector ${X^{ta}}_{ds}\in R^{N*C*\frac{H}{C}}$.
\begin{equation}
\left\{
             \begin{array}{lr}
             Q = FC(X_{ds}) \\
             K = FC(X_{ds}) \\
             V = FC(X_{ds}) ,
             \end{array}
\right.
\end{equation}
\begin{equation}
{X^{ta}}_{ds} = softmax(Q*K^T)V,
\end{equation}
where, $FC(.)$ stands for the fully connected layer. $softmax(.)$ stands for the normalized exponential function. $K^T$ stands for $K$ after converting the last two dimensions ($K^T \in R^{N*\frac{H}{C}*C}$).

Then, the output ${X^{TA}}_{ds}\in R^{N*C*\frac{H}{C}}$ of the temporal attention component is obtained by the residual connection  and the layer normalization:

\begin{equation}
U = \frac{1}{L}\sum_{i=0}^{L} ({X^{ta}}_{dsi} + X_{dsi}),
\end{equation}

\begin{equation}
\sigma = \sqrt{\frac{1}{L}\sum_{i=0}^{L} ({X^{ta}}_{dsi} + X_{dsi}-U)^2},
\end{equation}

\begin{equation}
{X^{TA}}_{ds} = \frac{g}{\sqrt{\sigma^2+\epsilon}} \odot ({X^{ta}}_{ds} + X_{ds}-U) + b,
\end{equation}
where, $U$ and $\sigma$ are represent the statistics of the feature vectors. $g$ is the gain. $b$ is the bias. $\sigma$ is a small decimal number that prevents division by zero.

\textbf{Variable attention.} Different from temporal attention, variable attention mainly uses multi-head attention to mine data from the perspective of the number $N$ of variables. Through the mining of variable attention, DSformer can effectively analyze the correlation between different variables and conduct information interaction. The formulas of the variable attention are given as follows:
\begin{equation}
\left\{
             \begin{array}{lr}
             Q = FC(X_{ds}) \\
             K = FC(X_{ds}) \\
             V = FC(X_{ds}), 
             \end{array}
\right.
\end{equation}
\begin{equation}
{X^{VA}}_{ds} = softmax(\frac{Q*K^T}{\sqrt{d_k}})V,
\end{equation}
where, $d_k$ can let the outcome of $Q*K^T$ satisfy the distribution with expectation 0 and variance 1. In particular, in the above formula, the corresponding matrix forms of $Q$ and $K^T$ are $\frac{H}{C}*C*N$ and $\frac{H}{C}*N*C$, respectively

Based on above temporal attention method and variable attention method, ${X^{TA}}_{ds}\in R^{N*C*\frac{H}{C}}$ and ${X^{VA}}_{ds}\in R^{N*C*\frac{H}{C}}$ are obtained. Then, ${X^{TA}}_{ds}$ and ${X^{VA}}_{ds}$ are integrated and the output ${X^{'}}_{ds}\in R^{N*\frac{H}{C}}$ of TVA block is obtained by the following formula:
\begin{equation}
{X^{'}}_{ds} = FC(F_{LN}({X^{TA}}_{ds}+{X^{VA}}_{ds})),
\end{equation}
where, $F_{LN}(.)$ stands for layer normalization. In addition, the main function of $FC(.)$ is to transform the feature vector dimension from $N*C*\frac{H}{C}$ to $N*\frac{H}{C}$.

\textbf{Information integration based on TVA block.} The feature vector $X_{ps}\in R^{N*C*\frac{H}{C}}$ is mined in the same way as above methods. And the output feature vector ${X^{'}}_{ps}\in R^{N*\frac{H}{C}}$ is obtained. For ${X^{'}}_{ps}$ and ${X^{'}}_{ds}$, we first used the layer normalization to achieve preliminary information fusion.
\begin{equation}
X^{'} = F_{LN}({X^{'}}_{ps}+{X^{'}}_{ds}),
\end{equation}
Then, the two-dimensional feature vector $X^{'}\in R^{N*\frac{H}{C}}$ was used as the input to the TVA block and fully mined from the temporal dimension and variable dimension.

Different from the previous modeling form, the feature vectors ${X^{'}}_{ps}\in R^{N*\frac{H}{C}}$ processed by the the information fusion method based on TVA block are 2D tensors. Therefore, the main matrix forms of the variables $Q$ and $K^T$ (temporal attention) modeled here are $N*\frac{H}{C}$ and $\frac{H}{C}*N$ respectively.

Finally, the feature vectors, which are further mined and integrated by TVA block, are passed to the multi-layer perceptron to effectively realize the output of the final prediction result $Y\in R^{N*L}$. 

\subsection{DSformer}
DSformer is constructed by effectively combining the double sampling block and three TVA block. The double sampling block effectively obtains the feature vectors containing key information. TVA block mines different feature vectors and fully realizes information integration. The specific modeling steps of the proposed DSformer are given as follows:

Step I: For the original 2D input features $X\in R^{N*H}$, the data is transformed into two 3D features $X_{ds}\in R^{N*C*\frac{H}{C}}$ and $X_{ps}\in R^{N*C*\frac{H}{C}}$ by a double sampling block.

Step II: Two TVA blocks are used to model and analyze $X_{ds}$ and $X_{ps}$, respectively. TVA block deeply mines feature vectors from both temporal dimension and variable dimension. In addition to this, the 3D features are transformed into 2D features ${X^{'}}_{ds}\in R^{N*\frac{H}{C}}$ and ${X^{'}}_{ps}\in R^{N*\frac{H}{C}}$ by the feedforward neural network.

Step III: Add ${X^{'}}_{ds}$ and ${X^{'}}_{ps}$. And then layer normalization is used to process the new feature vector $X^{'}\in R^{N*\frac{H}{C}}$.

Step IV: The TVA block is used to further mine the feature vector $X^{'}$ from the temporal dimension and the variable dimension. At the same time, the mined feature vectors are passed to the MLP for long-term prediction.

Step V: Based on the MLP for decoding, the model finally obtains the prediction result $Y\in R^{N*L}$ with the prediction step of $L$. The decoding process is calculated using the following formula:
\begin{equation}
Y = FC(X^{''}),
\end{equation}
where, $X^{''}$ stands for the feature vector obtained after the TVA block processing in step IV.

Based on the above steps, DSformer can effectively analyze and mine the key features and obtain the multivariate time series long-term prediction results. In addition, to ensure the training effect of the model, we adopt the method of fusing L1 Loss and L2 Loss for error backpropagation. The formula is given as follows:
\begin{equation}
\begin{aligned}
Loss = w_{L1} * \frac{1}{B*N*L}\sum_{k=0}^{B}\sum_{j=0}^{N}\sum_{i=0}^{L} |Y_{ijk}-{Y^{tru}}_{ijk}| + \\
 (1-w_{L1}) * \frac{1}{B*N*L}\sum_{k=0}^{B}\sum_{j=0}^{N}\sum_{i=0}^{L} (Y_{ijk}-{Y^{tru}}_{ijk})^2,
\end{aligned}
\end{equation}
where, $Y$ represents the prediction result of the model. $Y^{tru}$ stands for the true label. $B$ represents the number of samples. $N$ stands for the number of variables. $L$ stands for prediction step size. $w_{L1}$ represents the weight of Loss.

\section{Experiment and Analysis}
\subsection{Experimental design}
\textbf{Dataset.} In order to fully verify the effectiveness of the proposed DSformer in the field of multivariate time series long-term prediction, this paper selects nine classical data sets for comparative experiments. These datasets include ETT (ETTh1, ETTh2, ETTm1 and ETTm2), Exchange, ILI, Weather, Electricity and Traffic \cite{RN912}. Table 1 presents the basic statistics of these datasets.

\begin{table}[h]
  \caption{The statistics of the nine datasets.}
  \label{tab:freq}
  \begin{tabular}{cccc}
    \toprule
    Datasets&Variates &Timesteps&Granularity\\
    \midrule
ETTh1&7&17420&1hour\\
ETTh2&7&17420&1hour\\
ETTm1&7&69680&15min\\
ETTm2&7&69680&15min\\
Exchange&8&7588&1day\\
ILI&7&966&1week\\
Weather&21&52696&10min\\
Electricity&321&26304&1hour\\
Traffic&862&17544&1hour\\
  \bottomrule
\end{tabular}
\end{table}

\noindent\textbf{Baselines.} To construct comparative experiments and prove the effectiveness of DSformer, we select eight SOTA models with excellent performance in time series long-term prediction as baselines. The main baselines include PatchTST \cite{RN20}, Crossformer \cite{RN19}, TimesNet \cite{RN18}, Dlinear \cite{RN21}, FEDformer \cite{RN912}, Pyraformer \cite{RN909}, Autoformer \cite{RN886} and Informer \cite{RN742}. 

\noindent\textbf{Setting.} The main hyperparameter values of the DSformer are shown in Table 2. To conduct fair comparison experiments, we designed the experiment from the following aspects: 
(1) These nine datasets are divided into training sets, validation sets, and test sets according to the ratio in the reference \cite{RN50}. 
(2) These nine datasets were uniformly preprocessed by z-score normalization method. For each set of experiments, we set five different random seeds for repeated experiments. The final result of the model is obtained by averaging the repeated experiments.
(3) For the ILI dataset, we set the historical looking back window $H$ = 36 and the predicted future step size $L$ = [24, 36, 48, 60]. For the other data sets, we set the history looking back window $H$ = 96 and the prediction future step size $L$ = [96, 192, 336, 720]. 

\begin{table}[h]
  \caption{Values of the corresponding hyperparameters for different prediction step sizes.}
  \label{tab:freq}
  \begin{tabular}{cc}
    \toprule
    \multirow{2}*{Config}&Values\\
    &(96,192,336,720) \\
    \midrule
optimizer&Adam \cite{kingma2014adam} \\
learning rate&0.0001\\
number of multi-head attention&2/2/1/1\\
Dropout&0.15\\
sampling interval&2/2/3/3\\
weight of Loss&0.35/0.35/0.65/0.65\\
learning rate schedule&MultiStepLR\\
milestone&[25,50,75]\\
gamme&0.5\\
batch size&16\\
epoch&100\\
  \bottomrule
\end{tabular}
\end{table}

\noindent\textbf{Evaluation index.} The selection of appropriate evaluation indexes is the key to evaluate the prediction performance of different models. Considering the characteristics of multivariate long sequence time series prediction, we choose Mean Absolute Error (MAE) \cite{liu2021new} and Mean Squared Error (MSE)\cite{RN837} as the main evaluation indexes .

\subsection{Main results}
Table 3 shows the prediction results of the proposed DSformer and all baselines on nine datasets. The best results are highlighted in bold and the second best results are underlined. Based on Table 3, the following conclusions can be obtained: (1) Compared with other SOTA methods, Informer and Pyraformer have larger prediction errors. Although these two methods design advanced attention structures to improve the performance of the model, they fail to fully mine the core features of time series.
(2) Autoformer and FEDformer improve their prediction performance by introducing trend-season decomposition. However, the decomposition method converts the original sequence into a fixed form based on expert experience, which limits the ability of the model to mine the original data. 
(3) Compared with transformer variants mentioned above, Patch TST, TimesNet and Crossformer focus on mining global information and variable correlation from original data, which enables them to achieve better prediction results. However, for multivariate long sequence time series, they do not make full use of the three features proposed in this paper, which makes their performance limited.
(4) Compared with the existing SOTA models, the DSformer can achieve satisfactory prediction results. Firstly, two feature vectors focusing on global information and local information can be obtained through the DS block. Then, TVA block can mine and model these feature vectors from temporal dimension and variable dimension. Finally, DSformer uses a parallel structure to integrate the above feature information and realize long-term prediction. Therefore, the DSformer can achieve better performance than other SOTA methods.

\begin{table*}[h]
\footnotesize
  \caption{Multivariate time series prediction results on nine real-world datasets.}
  \label{tab:commands}
  \begin{tabular}{cccccccccccccccccccc}
    \toprule
    \multirow{2}*{Data} 
    & \multirow{2}*{$L$}  &\multicolumn{2}{c}{DSformer} 
    &\multicolumn{2}{c}{Patch TST $^{\mathrm{*}}$}&\multicolumn{2}{c}{Crossformer$^{\mathrm{*}}$}
    &\multicolumn{2}{c}{TimesNet}&\multicolumn{2}{c}{Dlinear}
    &\multicolumn{2}{c}{FEDformer} &\multicolumn{2}{c}{Autoformer}
    &\multicolumn{2}{c}{Informer} &\multicolumn{2}{c}{Pyraformer}
    \\
\cline{3-20} 
& & MSE & MAE & MSE & MAE & MSE & MAE & MSE & MAE& MSE & MAE& MSE & MAE& MSE & MAE& MSE & MAE& MSE & MAE\\

\midrule
 	 	 	 	 
\multirow{4}*{\rotatebox{90}{ETTh1}} 
&96&\textbf{0.352}&\textbf{0.392}
&0.393 &	0.408& 0.396 &	0.412& \underline{0.384} &	0.402&
0.386&\underline{0.400}&0.376&0.419&0.449&0.459&0.865&0.713&0.664&0.612\\

&192&\textbf{0.408}&\textbf{0.425}
&0.445 &	0.434&0.410 &	0.438& \underline{0.436} &	\underline{0.429}&0.437&0.432&0.420&0.448&0.500&0.482&1.008&0.792&0.790&0.681\\

&336&\underline{0.448}&\textbf{0.436}&0.484 &	\underline{0.451}&\textbf{0.440} &	0.461& 0.491 &	0.469&
0.481&0.459&0.459&0.465&0.521&0.496&1.107&0.809&0.891&0.738\\

&720&\textbf{0.469}&\textbf{0.454}&\underline{0.480} &	\underline{0.471}&0.519 &	0.524& 0.521& 	0.500 &
0.519&0.516&0.506&0.507&0.514&0.512&1.181&0.865&0.963&0.782\\

\midrule

\multirow{4}*{\rotatebox{90}{ETTh2}} 
&96&\textbf{0.268}&\textbf{0.304}&\underline{0.294} &\underline{0.343} 	&0.339 &	0.379 &	0.340 &	0.374&0.333&0.387&
0.346&0.388&0.358&0.397&3.489&1.515&0.645&0.597\\

&192&\textbf{0.332}&\textbf{0.341}&\underline{0.377} &	\underline{0.393}&0.415 &	0.425 	&0.402 &	0.414&0.477&0.476&
0.429&0.439&0.456&0.452&3.755&1.525&0.788&0.683\\

&336&\textbf{0.349}&\textbf{0.387}&\underline{0.381} &	\underline{0.409}& 0.452 &	0.468 &	0.452 &	0.452&0.594&0.541&
0.496&0.487&0.482&0.486&4.721&1.835&0.907&0.747\\

&720&\textbf{0.375}&\textbf{0.393}&\underline{0.412} &	\underline{0.433}&0.455 &	0.471 &0.462 &	0.468 &0.831&0.657& 
0.463&0.474&0.515&0.511&3.647&1.625&0.963&0.783\\

 \midrule
\multirow{4}*{\rotatebox{90}{ETTm1}} 
&96&\textbf{0.292}&\underline{0.368}&0.321 &\textbf{0.360}&\underline{0.320} &	0.373 &0.338 &	0.375&
0.345&0.372& 	
0.379&0.419&0.505&0.475&0.672&0.571&0.543&0.510\\

&192&\textbf{0.351}&\textbf{0.379}&\underline{0.362}&\underline{0.384}&	0.386 &	0.401& 0.374 &	0.387&
0.380&0.389&
0.426&0.441&0.553&0.496&0.795&0.669&0.557&0.537\\

&336&\textbf{0.384}&\underline{0.408}&\underline{0.392} &\textbf{0.402}&0.404 &	0.427 &	0.410 &	0.411 & 
0.413&0.413&	 
0.445&0.459&0.621&0.537&1.212&0.871&0.754&0.655\\

&720&\textbf{0.442}&\underline{0.439}&\underline{0.450} &\textbf{0.435}&0.569 &	0.528& 0.478 &	0.450&
0.474&0.453&
0.543&0.490&0.671&0.561&1.166&0.823&0.908&0.724\\

	 	 	 
\midrule
\multirow{4}*{\rotatebox{90}{ETTm2}} 
&96&\textbf{0.130}&\textbf{0.231}&\underline{0.178} &	\underline{0.260}&0.196 &	0.275& 0.187 &	0.267 &
0.193&0.292&
0.203&0.287&0.255&0.339&0.365&0.453&0.435&0.507\\

&192&\textbf{0.207}&\textbf{0.275}&\underline{0.249} &	\underline{0.307}&0.248 &	0.317 & 0.249& 0.309&	
0.284&0.362& 	
0.269&0.328&0.281&0.340&0.533&0.563&0.730&0.673\\

&336&\textbf{0.262}&\textbf{0.318}&\underline{0.313} &	\underline{0.346}& 0.322 &	0.358&	0.321& 	0.351 &
0.369&0.427&	
0.325&0.366&0.339&0.372&1.363&0.887&1.201&0.845\\

&720&\textbf{0.327}&\textbf{0.372}&\underline{0.400} &	\underline{0.398}&0.402 &	0.406& 0.408 &	0.403&
0.554&0.522&
0.421&0.415&0.433&0.432&3.379&1.338&3.625&1.451\\


\midrule
\multirow{4}*{\rotatebox{90}{Exchange}} 
&96&\textbf{0.075}&\textbf{0.213}&\underline{0.081} &	\underline{0.216}& 	 0.139 &	0.265& 0.107 &	0.234&
0.088&0.218&0.148&0.278&0.197&0.323&0.847&0.752&0.376&1.105\\

&192&\textbf{0.158}&\textbf{0.308}&\underline{0.169} &	\underline{0.317}& 	0.241 &	0.375& 0.226 &	0.344&
0.176&0.315&0.271&0.380&0.300&0.369&1.204&0.895&1.748&1.151\\

&336&\textbf{0.294}&\textbf{0.402}&\underline{0.305} &	\underline{0.416}& 0.392 	&0.468& 0.367 &	0.448&
0.313&0.427&0.460&0.500&0.509&0.524&1.672&1.036&1.874&1.172\\

&720&\textbf{0.834}&\textbf{0.692}&	 0.853 &	0.702&1.112 &	0.802 &0.964 &	0.746&
\underline{0.839}&\underline{0.695}&1.195&0.841&1.447&0.941&2.478&1.310&1.943&1.206\\


\midrule
\multirow{4}*{\rotatebox{90}{ILI}} 
&24&\textbf{1.538}&\underline{0.815}&\underline{1.610} &\textbf{0.814}&  3.041 &1.186& 2.317 &	0.934&
2.398&1.040&3.228&1.260&3.483&1.287&5.764&1.677&7.042&2.012\\

&36&\textbf{1.546}&\textbf{0.829}&\underline{1.579}&\underline{0.870}& 	3.406 &	1.232& 1.972 &	0.920&
2.646&1.088&2.679&1.080&3.103&1.148&4.755&1.467&7.394&2.031\\

&48&\textbf{1.672}&\textbf{0.841}&\underline{1.673}&\underline{0.854}& 3.459 & 1.221 & 2.238 &	0.940&
2.614&1.086&2.622&1.078&2.669&1.085&4.763&1.469&7.551&2.057\\

&60&\textbf{1.548}&\textbf{0.803}&\underline{1.702}&\underline{0.829}&3.640 & 1.305 &2.027 &	0.928&
2.804&1.146&2.857&1.157&2.770&1.125&5.264&1.564&7.662&2.100\\

 	 	 	 
\midrule
\multirow{4}*{\rotatebox{90}{Weather}} 
&96&\textbf{0.170}&\textbf{0.217}&0.178&\underline{0.219}&  0.185 &	0.248& \underline{0.172} &	0.220&	
0.196&0.255&0.217&0.296&0.266&0.336&0.300&0.384&0.896&0.556\\

&192&\textbf{0.215}&\textbf{0.257}&0.224&\underline{0.259}& 0.229 &	0.305 &\underline{0.219} &	0.261&
0.237&0.296&0.276&0.336&0.307&0.367&0.598&0.544&0.622&0.624\\

&336&\textbf{0.265}&\textbf{0.295}&\underline{0.278}&\underline{0.298}& 0.287 &	0.332& 0.280 &	0.306&
0.283&0.335&0.339&0.380&0.359&0.395&0.578&0.523&0.739&0.753\\

&720&\textbf{0.322}&\textbf{0.342}&\underline{0.350}&\underline{0.346}&0.356 &	0.398 & 0.365 &	0.359& 
0.345&0.381&0.403&0.428&0.419&0.428&1.059&0.741&1.004&0.934\\

\midrule
\multirow{4}*{\rotatebox{90}{Electricity}} 
&96&\textbf{ 0.163}&\underline{0.264}&0.174 &\textbf{0.259}& 	 0.175 &	0.279& \underline{0.168} &	0.272&
0.197&0.282&0.193 &0.308& 0.201 &0.317 & 0.274 &0.368& 0.386&0.449 \\

&192&\textbf{0.174}&\underline{0.272}&\underline{0.178} &\textbf{0.265}& 0.192 &	0.302&	0.184 &	0.289&
0.196 &	0.285&0.201 &0.315& 0.222& 0.334& 0.296 &0.386& 0.378&0.443\\

&336&\textbf{0.187}&\underline{0.287}&\underline{0.196}&\textbf{0.282}&	0.208 &	0.317 & 0.198 &	0.300&
0.196 &	0.285&0.201 &0.315& 0.222 &0.334& 0.296 &0.386&0.376&0.443\\

&720&\textbf{0.216}&\textbf{0.309}&0.237 &	\underline{0.316}&0.225 &	0.337& \underline{0.220} &	0.320&
0.245&0.333&0.246&0.355&0.254 &0.361& 0.373&0.439& 0.376 &0.445\\

 	 	 	 
\midrule
\multirow{4}*{\rotatebox{90}{Traffic}} 
&96&\textbf{0.458}&0.311&\underline{0.477} &\underline{0.305}&  0.519 &\textbf{0.295}&0.593 &	0.321&
0.650&0.396&0.587&0.366&0.613&	0.388&0.719&0.391&0.867 &0.468\\

&192&\textbf{0.467}&0.323&\underline{0.471} &\textbf{0.299}&0.526 &	\underline{0.307} &	0.617 &	0.336&
0.598 &	0.370&0.604 &0.373 &0.616&0.379 & 0.696 &0.382& 0.869 &0.467\\

&336&\textbf{0.479}&0.329&\underline{0.485} &\underline{0.305}& 0.530 &\textbf{0.300}&0.629 &0.336 & 
0.605&0.373&0.621&0.383& 0.622&0.337&0.777 &0.420 &0.881 &0.469\\

&720&\textbf{0.512}&0.342&\underline{0.518}&\underline{0.325}&0.573 &	\textbf{0.313} &0.640 &	0.350&
0.645&0.394 &0.626 &0.382&0.660&0.408&0.864&0.472&0.896&0.473\\

    \bottomrule

\multicolumn{20}{l}{$^{\mathrm{*}}$ represents that we set uniform input and output sizes to ensure the fairness of the experiment. The results of other methods are from Timesnet \cite{RN18}}
    
  \end{tabular}
\end{table*}

\subsection{Ablation experiments}
The DSformer contains four key components: piecewise sampling, down sampling, temporal attention and variable attention. To prove that these components can help the DSformer to mine the key feature information, wu conducts ablation experiments from the following five perspectives:  (1) wo/ ps: the piecewise sampling component is removed. (2) wo/ ds: the down-sampling component is removed. (3) wo/ as: In this part, we remove the double sampling block. (4) wo/ ta: we remove temporal attention and replace it with multi-layer perceptron.
(5) wo/ va: variable attention is removed. 

\begin{figure}[h]
  \centering
  \includegraphics[width=\linewidth]{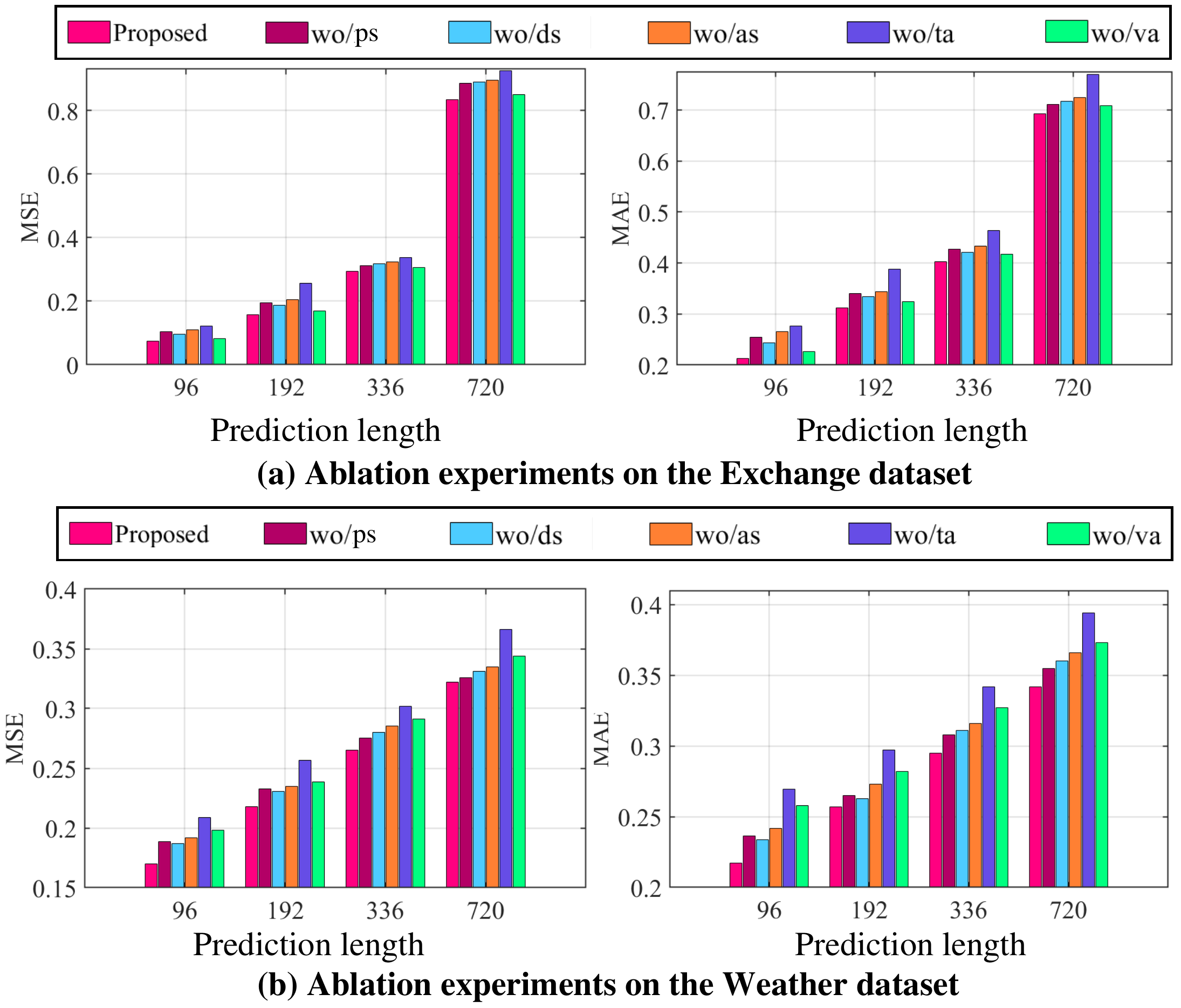}
  \caption{Results of ablation experiments on Exchange and weather datasets.}
  \Description{Results of ablation experiments.}
\end{figure}

Figure 5 illustrates the results of the ablation experiments. Based on the experimental results, we can obtain the following conclusions: (1) When there is a correlation between different variables, deleting the variable attention will increase the prediction error of the model. According to the experimental results, the correlation between different time series in Weather data set is large, so the variable attention have a great influence on the prediction results. (2) After deleting the temporal attention, the prediction performance of DSformer decreases significantly. The main reason is that the most important step in time series modeling is mining the time series context relation. If temporal attention is lost, it is difficult for DSformer to effectively analysis the context relation of time series data. (3) When the prediction step size is long, removing down sampling significantly increases the error of the prediction. When the prediction step size is short, deleting piecewise sampling significantly increases the prediction error. The main reason is that the features obtained by down sampling contain more global information, and the features obtained by interval sampling contain more local information. Therefore, they will affect the modeling effect of different prediction steps, respectively. (4) After removing the down-sampling and piecewise sampling at the same time, the prediction error of the DSformer further increases. The main reason is that these two sampling methods can deepen the model's ability to focus on learning the global and the local respectively. When the sampling method is removed, the model may not be able to focus on the key information, which increases the prediction errors of the DSformer. (5) The results of ablation experiments can demonstrate the importance of the proposed four components, which can effectively mine the three main features of multivariate long sequence time series and reduce prediction error.

\subsection{Hyperparameter analysis experiments}
The hyperparameters of the DSformer usually affect the final prediction results. In order to fully analyze the sensitivity of the DSformer and the role of some key hyperparameters, this section conducts an experimental analysis of the four main hyperparameters (learning rate, number of multi-head attention, sampling interval and weight of Loss) on ETTm2 dataset. Figure 6 illustrates the results of the hyperparameter analysis experiments. 

\begin{figure}[h]
  \centering
  \includegraphics[width=\linewidth]{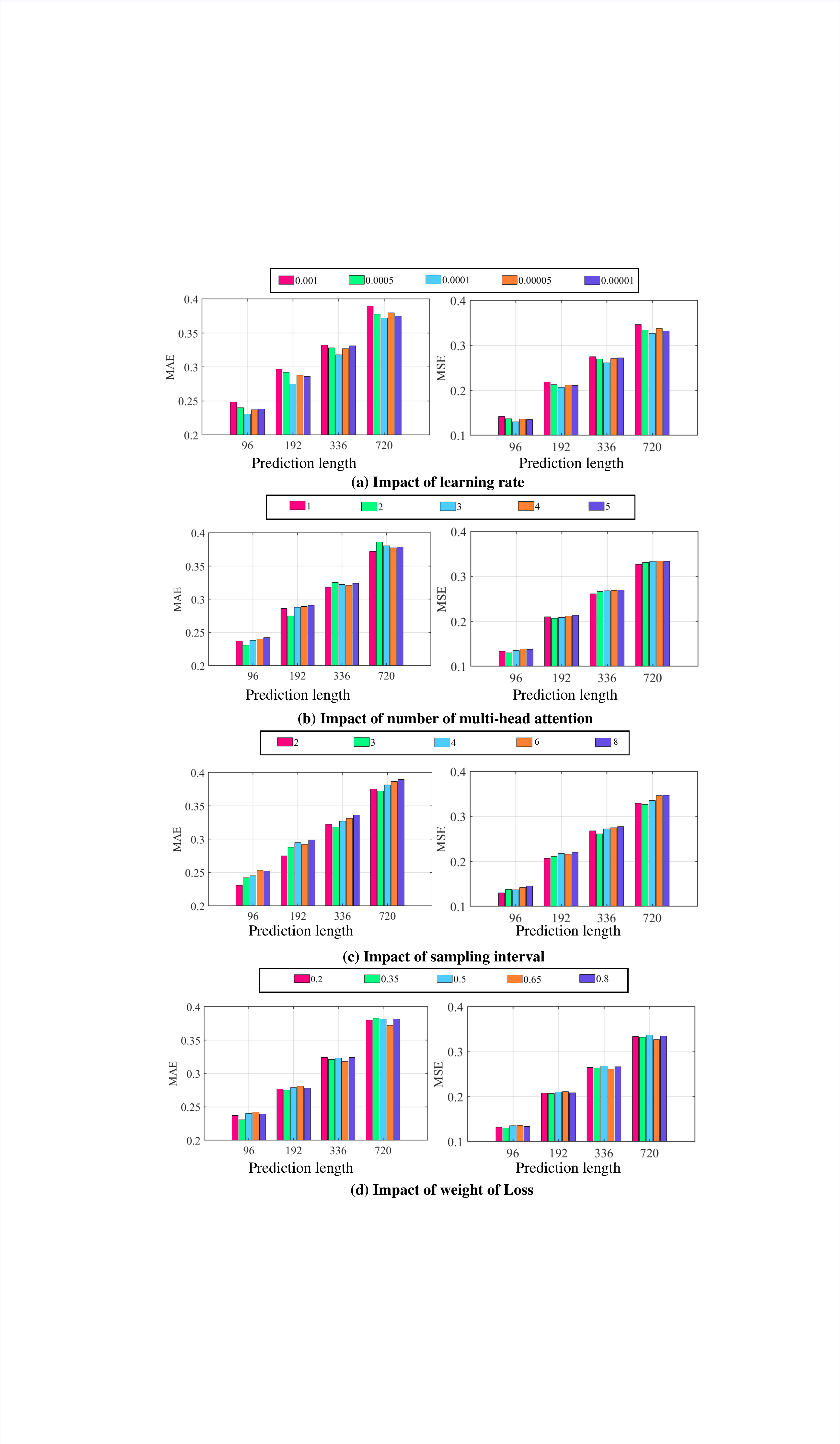}
  \caption{Impact of four key hyperparameters on prediction results (ETTm2 dataset).}
  \Description{Impact of four key hyperparameters on experimental results.}
\end{figure}

Based on the experimental results, we can get the following conclusions: (1) The number of multi-head attention and the weight of Loss have relatively little impact on the prediction performance of the model. For multi-head attention, an appropriate number can prevent overfitting while ensuring modeling performance. For the weight of Loss, an appropriate value can ensure the training effect of the model and improve the overall performance. (2) The learning rate has a large impact on the model performance. The main reasons include the following two aspects: On the one hand, a large learning rate will produce phenomena such as overfitting. On the other hand, a smaller learning rate will affect the effect of training. Therefore, the setting of learning rate is very important to ensure the training effect of DSformer. (3) As one of the main hyperparameters, the sampling interval has a relatively large impact on the prediction results. When the sampling interval is small, DSformer can achieve better results with shorter prediction steps. When the sampling interval is relatively large, DSformer can achieve better results on longer prediction steps. However, when the sampling interval is too large, the prediction error of DSformer increases significantly. The main reason is that too large sampling interval makes the model lose more local information, which resulted in insufficient usage of information and reduced prediction accuracy. Therefore, setting the sampling interval appropriately can affect the effect of different prediction steps of the DSformer.(4) The sampling interval $C$ of the double sampling block is an important parameter because it affects the input feature information of the DSformer. In the next section, we will future analyze the influence of sampling interval $C$ and input history length $H$ on the experimental results.

\subsection{Effects of the sampling interval and the history length}
The history length affects the information obtained by the model. And the sampling interval affects the model's utilization of information. Considering the effect of the history length and the sampling interval on the input information of DSformer, we further analyze the influence of these two hyperparameters on the prediction results in this section. To adequately evaluate these two key hyperparameters, we carried out the following comparison experiments: (1) Based on the hyperparameter experiments, it can be found that too large sampling interval is not conducive to the experimental results. Therefore, we set the sampling interval of the double sampling block, that is, the number of subsequences $C = [2,3,4,6]$. (2) Considering the characteristics of the prediction step size of the model, the history length of the model is set as $H = [96,192,336]$, respectively. (3) All above parameters were used to construct experiments on the ETTh2 dataset. And the future length of the DSformer is set to $L = [96,192,336,720]$.

\begin{table}[h]
  \caption{Experimental results on ETTh2 dataset with different history length $H$ and sampling interval $C$.}
  \label{tab:commands}
  \begin{tabular}{cccccccc}
    \toprule
    \multirow{2}*{$L$} & $H$&\multicolumn{2}{c}{96} &\multicolumn{2}{c}{192} &\multicolumn{2}{c}{336} \\
\cline{2-8} 
&$C$& MSE & MAE & MSE & MAE & MSE & MAE\\
\midrule
\multirow{4}*{96} 
&2&0.268 &	0.304 &	0.269 &	0.313 &	0.289 &	0.342\\
&3&0.273 &	0.311 &	0.278 &	0.328 &	0.274 &	0.325\\
&4&0.275 &	0.315 &	0.263 &	0.305 &	0.254 &	0.295\\
&6&0.278 &	0.327 &	0.265 &	0.307 &	0.262 &	0.302 \\

 \midrule
\multirow{4}*{192} 
&2&0.332 &	0.341 &	0.340 &	0.351 &	0.356 &	0.364 \\
&3&0.337 &	0.347 &	0.329 &	0.340 &	0.342 &	0.352 \\
&4&0.343 &	0.349 &	0.328 &	0.335 &	0.318 &	0.329 \\
&6&0.344 &	0.356 &	0.327 &	0.336 &	0.323 &	0.331 \\

\midrule
\multirow{4}*{336} 
&2&0.352 &	0.391 &	0.354 &	0.392 &	0.357 &	0.395\\
&3&0.349 &	0.387 &	0.352 &	0.393 &	0.353 &	0.391\\
&4&0.356 &	0.398 &	0.345 &	0.384 &	0.341 &	0.378\\
&6&0.362 &	0.401 &	0.347 &	0.386 &	0.337 &	0.376\\

\midrule
\multirow{4}*{720} 
&2&0.381 &	0.398 &	0.379 &	0.397 &	0.377 &	0.395\\
&3&0.375 &	0.393 &	0.374 &	0.392 &	0.371 &	0.392\\
&4&0.389 &	0.403 &	0.367 &	0.385 &	0.366 &	0.389\\
&6&0.394 &	0.412 &	0.365 &	0.386 &	0.362 &	0.383\\

    \bottomrule
  \end{tabular}
\end{table}

Table 4 shows the experimental results for different history lengths and sampling intervals. Based on the experimental results, the following conclusions can be drawn: 
(1) When the history length is short, the sampling interval $C$ cannot be too large. If the sampling interval is large, the prediction performance of the model will degrade significantly. The main reason is that the increasing sampling interval limits the ability of DSformer to mine the local information of raw data. The loss of too much local information is not conducive to the short-term prediction effect of the DSformer.
(2) When the history length is large, increasing the sampling interval $C$ can improve the prediction performance of the model. On the one hand, increasing the sampling interval can make the feature vector obtained by down-sampling contain more global information. On the other hand, when the history length is large, the model contains more historical period information, and increasing the sampling interval can make the piecewise sampling obtain feature vectors that pay more attention to local information.
(3) The history length $H$ of DSformer can affect the amount of global and local information obtained by the model. The size of the sampling interval $C$ can make the model focus on different informations of input features. Therefore, the appropriate balance between the history length and the sampling interval can make the model effectively use the overall information and the global information, which can improve the prediction accuracy of DSformer.

\subsection{Efficiency}
In this section, based on the Electricity data sets, we compare the efficiency of the DSformer and other baselines (Dlinear, Pyraformer, Crossformer, FEDformer and Autoformer).  Besides, to make a fair comparison, we compare the mean training time of each epoch of several models. The experimental equipment is the Intel(R) Xeon(R) Gold 5217 CPU @ 3.00GHz, 128G RAM computing server with RTX 3090 graphics card. The batch size is set to 16.

\begin{figure}[h]
  \centering
  \includegraphics[width=\linewidth]{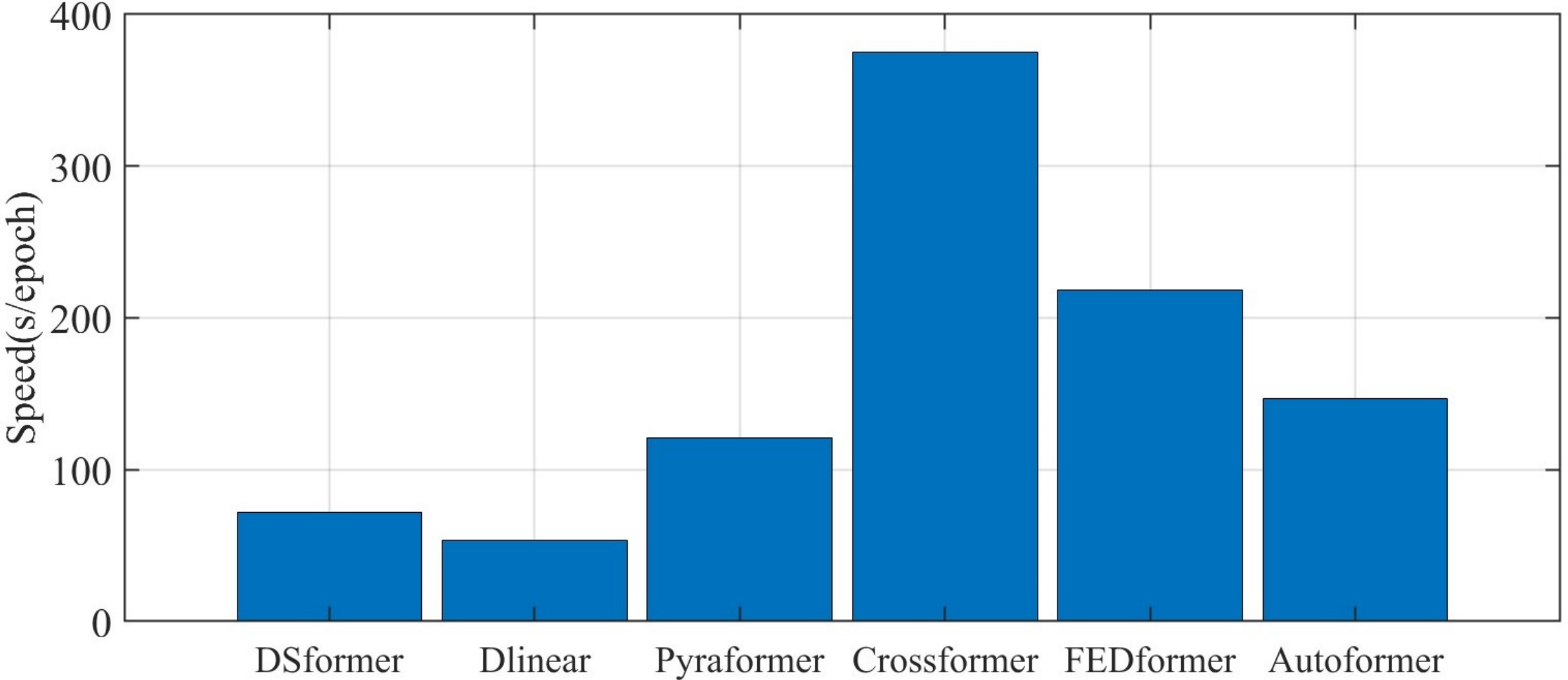}
  \caption{Training time for each epoch of different models.}
  \Description{Training time for each epoch of different models.}
\end{figure}

Based on Figure 7, it can be found that although the computational complexity of DSformer is $O(N^2)$, the actual computational resource consumption of DSformer is not large. Specifically, most existing transformer variants use various theoretical methods to reduce computational complexity, but their actual computational resource consumption is not low due to the introduction of many tricks. Compared with above models, DSformer has two advantages: On the one hand, DSformer uses the DS block to reduce the length of the sequence that  needs to be modeled. On the other hand, DSformer does not use some tricks that significantly increase computational resource consumption, such as embeding. Therefore, the results of efficiency comparison further prove the practical application value of DSformer.

\section{Conclusion and Future Work}
In this paper, we propose DSformer, an efficient multivariate time series long-term prediction model, which contains two finely designed blocks, including the DS block and TVA block. The DS block simply and efficiently mines the global information and the local information of time series, which are significant features for long-term prediction. And the TVA block can effectively integrate the above information and variable correlation to significantly improve time series prediction accuracy. The experiments on nine real-world datasets show that DSformer achieves state-of-the-art performance for MTS long-term prediction. In the future, we will try to design a module to adaptive balance sampling interval and history length, further improving the information mining ability and long term prediction effect of the model.

\begin{acks}
This work is supported by the Youth Innovation Promotion Association CAS No.2023112.
\end{acks}

\clearpage

\bibliographystyle{ACM-Reference-Format}
\bibliography{sample-base}


\end{document}